\documentclass[conference]{IEEEtran}
\IEEEoverridecommandlockouts
\usepackage{cite}
\usepackage{amsmath,amssymb,amsfonts}
\usepackage{graphicx}
\usepackage{textcomp}
\usepackage{xcolor}
\usepackage{booktabs}
\usepackage{balance}
\usepackage{url}
\graphicspath{{figs/}}

\newcommand{\GridCells}{360}

\newcommand{\GridSgdmFloor}{\ensuremath{-}21.2}

\newcommand{\GridSgdmNegCells}{29}
\newcommand{\GridAdamWorst}{\ensuremath{-}46.5}
\newcommand{\GridAdamNegCells}{174}
\newcommand{\GridAdamNegPct}{48}

\newcommand{\GridConfigs}{72}
\newcommand{\GridUnanimousConfigs}{69}

\newcommand{\GridWarmGridLo}{200}
\newcommand{\GridWarmCapStep}{20,000}
\newcommand{\GridWarmGridCount}{7}
\newcommand{\GridWarmCapCells}{42}
\newcommand{\GridPThreeSgdmWinMean}{1.03}
\newcommand{\GridPThreeAdamWinMean}{1.37}

\newcommand{\GridPTwoGap}{+0.01}

\newcommand{\GridPOneGap}{\ensuremath{-}0.03}

\newcommand{\GridLNineSixAdamNegPct}{36}

\newcommand{\GridLOneNineTwoAdamNegPct}{61}

\newcommand{\LrCells}{360}

\newcommand{\LrLNineSixFixedSgdmWins}{138}
\newcommand{\LrLNineSixFixedAdamWins}{42}
\newcommand{\LrLNineSixFixedSgdmNegCells}{8}

\newcommand{\LrLNineSixFixedSgdmMin}{\ensuremath{-}4.4}
\newcommand{\LrLNineSixFixedAdamNegCells}{65}

\newcommand{\LrLNineSixFixedAdamMin}{\ensuremath{-}29.0}

\newcommand{\LrLNineSixFixedMedianGapPt}{\ensuremath{-}5.6}
\newcommand{\LrLNineSixFixedCfgAdamWins}{9}
\newcommand{\LrLNineSixFixedCfgSgdmWins}{27}

\newcommand{\LrLNineSixSelSgdmWins}{23}
\newcommand{\LrLNineSixSelAdamWins}{157}
\newcommand{\LrLNineSixSelSgdmNegCells}{6}

\newcommand{\LrLNineSixSelSgdmMin}{\ensuremath{-}4.6}
\newcommand{\LrLNineSixSelAdamNegCells}{4}

\newcommand{\LrLNineSixSelAdamMin}{\ensuremath{-}1.3}

\newcommand{\LrLNineSixSelMedianGapPt}{+2.1}
\newcommand{\LrLNineSixSelCfgAdamWins}{32}
\newcommand{\LrLNineSixSelCfgSgdmWins}{4}

\newcommand{\LrLNineSixOrcSgdmWins}{2}
\newcommand{\LrLNineSixOrcAdamWins}{178}
\newcommand{\LrLNineSixOrcSgdmNegCells}{0}

\newcommand{\LrLNineSixOrcSgdmMin}{+0.1}
\newcommand{\LrLNineSixOrcAdamNegCells}{0}

\newcommand{\LrLNineSixOrcAdamMin}{+0.1}

\newcommand{\LrLNineSixOrcMedianGapPt}{+1.3}
\newcommand{\LrLNineSixOrcCfgAdamWins}{36}
\newcommand{\LrLNineSixOrcCfgSgdmWins}{0}

\newcommand{\LrLOneNineTwoFixedSgdmWins}{177}
\newcommand{\LrLOneNineTwoFixedAdamWins}{3}
\newcommand{\LrLOneNineTwoFixedSgdmNegCells}{21}

\newcommand{\LrLOneNineTwoFixedSgdmMin}{\ensuremath{-}21.2}
\newcommand{\LrLOneNineTwoFixedAdamNegCells}{109}

\newcommand{\LrLOneNineTwoFixedAdamMin}{\ensuremath{-}46.5}

\newcommand{\LrLOneNineTwoFixedMedianGapPt}{\ensuremath{-}12.8}
\newcommand{\LrLOneNineTwoFixedCfgAdamWins}{1}
\newcommand{\LrLOneNineTwoFixedCfgSgdmWins}{35}

\newcommand{\LrLOneNineTwoSelSgdmWins}{27}
\newcommand{\LrLOneNineTwoSelAdamWins}{153}
\newcommand{\LrLOneNineTwoSelSgdmNegCells}{1}

\newcommand{\LrLOneNineTwoSelSgdmMin}{\ensuremath{-}0.5}
\newcommand{\LrLOneNineTwoSelAdamNegCells}{0}

\newcommand{\LrLOneNineTwoSelAdamMin}{+0.3}

\newcommand{\LrLOneNineTwoSelMedianGapPt}{+1.0}
\newcommand{\LrLOneNineTwoSelCfgAdamWins}{32}
\newcommand{\LrLOneNineTwoSelCfgSgdmWins}{4}

\newcommand{\LrLOneNineTwoOrcSgdmWins}{6}
\newcommand{\LrLOneNineTwoOrcAdamWins}{174}
\newcommand{\LrLOneNineTwoOrcSgdmNegCells}{0}

\newcommand{\LrLOneNineTwoOrcSgdmMin}{+0.4}
\newcommand{\LrLOneNineTwoOrcAdamNegCells}{0}

\newcommand{\LrLOneNineTwoOrcAdamMin}{+0.4}

\newcommand{\LrLOneNineTwoOrcMedianGapPt}{+1.1}
\newcommand{\LrLOneNineTwoOrcCfgAdamWins}{35}
\newcommand{\LrLOneNineTwoOrcCfgSgdmWins}{1}

\newcommand{\LrAdamAtOneEMinusFourNegCells}{2}
\newcommand{\LrAdamAtOneEMinusFourMean}{+15.1}

\newcommand{\LrAdamWinsAtOneEMinusFour}{322}

\newcommand{\LrSgdmAtThreeEMinusFourMean}{+12.1}

\newcommand{\LrSgdmAtOneEMinusThreeNegCells}{29}

\newcommand{\LrAdamAtOneEMinusThreeNegCells}{174}

\newcommand{\LrAdamWinsAtOneEMinusThree}{45}

\newcommand{\LrAdamSelLeqThreeEMinusFourCells}{340}
\newcommand{\LrAdamSelGeqOneEMinusThreeCells}{20}

\newcommand{\LrSgdmSelNegCellsAll}{7}

\newcommand{\LrSgdmSelMinAll}{\ensuremath{-}4.6}
\newcommand{\LrSelAdamNegCellsAll}{4}
\newcommand{\LrSelAdamMinAll}{\ensuremath{-}1.3}
\newcommand{\LrSgdmSelExtCells}{14}

\newcommand{\LrSgdmSelNegAppliancesDlinearCells}{1}
\newcommand{\LrTolGuardChangedCells}{337}
\newcommand{\LrTolGuardMedianCostPt}{3.6}
\newcommand{\LrSelAdamVsOrcSgdmWins}{246}
\newcommand{\LrSelAdamVsOrcSgdmMedianPt}{+0.38}
\newcommand{\LrSelCfgMeanGapMedianPt}{+1.4}
\newcommand{\LrSelCfgMeanGapCiLoPt}{+1.0}
\newcommand{\LrSelCfgMeanGapCiHiPt}{+2.7}
\newcommand{\LrSelCfgMeanGapPosCfgs}{61}
\newcommand{\LrSelCfgMeanGapWilcoxonP}{1{\times}10^{-9}}
\newcommand{\LrSelDsBbMeanGapPos}{11}
\newcommand{\LrSelDsBbTotal}{12}
\newcommand{\LrAdamSelTenthToThirtiethCells}{268}

\newcommand{\LrSelAdamWinsAll}{310}

\newcommand{\LrSelMedianGapPtAll}{+1.5}

\newcommand{\LrConfigs}{72}
\newcommand{\LrHTwoFourCells}{120}
\newcommand{\LrHTwoFourSelAdamWins}{100}

\newcommand{\LrHFourEightSelAdamWins}{102}

\newcommand{\LrHNineSixSelAdamWins}{108}

\newcommand{\MfiveBdgTwoDlinearSelBestMean}{+6.4}

\newcommand{\MfiveBdgTwoPatchtstSelBestMean}{+13.5}
\newcommand{\MfiveBdgTwoPatchtstSelBestMin}{+12.0}
\newcommand{\MfiveBdgTwoPatchtstSelBestMax}{+15.0}
\newcommand{\MfiveBdgTwoFleetDlinearSelBestMean}{+7.0}

\newcommand{\MfiveBdgTwoFleetPatchtstSelBestMean}{+28.7}
\newcommand{\MfiveBdgTwoFleetPatchtstSelBestMin}{+25.8}
\newcommand{\MfiveBdgTwoFleetPatchtstSelBestMax}{+31.9}
\newcommand{\MfiveBdgTwoFoxDlinearSelBestMean}{+7.5}

\newcommand{\MfiveBdgTwoFoxPatchtstSelBestMean}{+11.4}
\newcommand{\MfiveBdgTwoFoxPatchtstSelBestMin}{+9.9}
\newcommand{\MfiveBdgTwoFoxPatchtstSelBestMax}{+13.8}
\newcommand{\MfiveBdgTwoPantherDlinearSelBestMean}{+4.8}

\newcommand{\MfiveBdgTwoPantherPatchtstSelBestMean}{+5.6}
\newcommand{\MfiveBdgTwoPantherPatchtstSelBestMin}{+4.3}
\newcommand{\MfiveBdgTwoPantherPatchtstSelBestMax}{+7.3}
\newcommand{\MfiveBdgTwoRatAllDlinearSelBestMean}{+2.0}

\newcommand{\MfiveBdgTwoRatAllPatchtstSelBestMean}{+43.7}
\newcommand{\MfiveBdgTwoRatAllPatchtstSelBestMin}{+39.8}
\newcommand{\MfiveBdgTwoRatAllPatchtstSelBestMax}{+47.1}
\newcommand{\MfiveBdgTwoRatWorstDlinearSelBestMean}{+2.1}

\newcommand{\MfiveBdgTwoRatWorstPatchtstSelBestMean}{+42.8}
\newcommand{\MfiveBdgTwoRatWorstPatchtstSelBestMin}{+41.6}
\newcommand{\MfiveBdgTwoRatWorstPatchtstSelBestMax}{+44.0}

\newcommand{\FroEttmTwoPatchtstCalibSgdMBenefit}{+8.2}
\newcommand{\FroEttmTwoPatchtstCalibSgdMBenefitStd}{5.3}

\newcommand{\FroEttmTwoPatchtstCalibSgdMParams}{16,934}
\newcommand{\FroEttmTwoPatchtstCalibSgdMMs}{2.88}

\newcommand{\FroEttmTwoPatchtstFullAdamBenefitFixed}{\ensuremath{-}19.6}

\newcommand{\FroEttmTwoPatchtstFullSgdMBenefit}{+10.2}
\newcommand{\FroEttmTwoPatchtstFullSgdMBenefitStd}{4.5}

\newcommand{\FroEttmTwoPatchtstFullSgdMMemKb}{685}
\newcommand{\FroEttmTwoPatchtstHeadAdamBenefit}{+10.7}

\newcommand{\FroEttmTwoPatchtstHeadAdamMemKb}{203}

\newcommand{\FroEttmTwoPatchtstHeadSgdMMs}{1.86}

\newcommand{\FroAppliancesPatchtstCalibAdamBenefit}{+44.5}

\newcommand{\FroAppliancesPatchtstCalibAdamMemKb}{204}

\newcommand{\FroAppliancesPatchtstCalibSgdMBenefit}{+42.1}

\newcommand{\FroAppliancesPatchtstCalibSgdMMemKb}{136}
\newcommand{\FroAppliancesPatchtstFullAdamBenefit}{+52.5}

\newcommand{\FroAppliancesPatchtstFullAdamMemKb}{1,028}

\newcommand{\FroAppliancesPatchtstFullSgdMBenefit}{+47.3}

\newcommand{\FroAppliancesPatchtstFullSgdMBenefitFixed}{+53.1}

\newcommand{\FroAppliancesPatchtstFullSgdMMemKb}{686}

\newcommand{\FroAppliancesPatchtstHeadAdamMemKb}{203}

\newcommand{\FroAppliancesPatchtstHeadSgdMMemKb}{135}
\newcommand{\FroEttmTwoFullOverCalibParams}{5.1}

\newcommand{\FroEnergyMinMj}{4.3}
\newcommand{\FroEnergyMaxMj}{18.5}
\newcommand{\FroMsMin}{0.85}
\newcommand{\FroMsMax}{3.69}
\newcommand{\FroSecPerYearMin}{0.31}
\newcommand{\FroSecPerYearMax}{1.35}

\newcommand{\StalEtthTwoWinPct}{\ensuremath{-}1.1}
\newcommand{\StalEtthTwoWinPctStd}{1.0}

\newcommand{\StalEttmTwoWinPct}{+2.6}
\newcommand{\StalEttmTwoWinPctStd}{1.4}

\newcommand{\StalAppliancesWinPct}{+5.6}
\newcommand{\StalAppliancesWinPctStd}{1.0}

\newcommand{\StalAdamEtthTwoWinPct}{\ensuremath{-}2.8}
\newcommand{\StalAdamEtthTwoWinPctStd}{1.6}
\newcommand{\StalAdamEttmTwoWinPct}{+4.5}
\newcommand{\StalAdamEttmTwoWinPctStd}{0.4}
\newcommand{\StalAdamAppliancesWinPct}{+1.9}
\newcommand{\StalAdamAppliancesWinPctStd}{0.3}

\newcommand{\WcMilestoneLo}{50}
\newcommand{\WcMilestoneHi}{50,000}
\newcommand{\WcMilestoneCount}{10}
\newcommand{\WcEttmTwoDlinearUnder}{+21.0}
\newcommand{\WcEttmTwoDlinearSweet}{\ensuremath{-}0.2}
\newcommand{\WcEttmTwoDlinearOver}{+4.6}
\newcommand{\WcEttmTwoDlinearSweetStep}{20,000}
\newcommand{\WcEttmTwoDlinearUnderStd}{6.5}
\newcommand{\WcEttmTwoDlinearSweetStd}{1.1}
\newcommand{\WcEttmTwoDlinearOverStd}{5.1}

\newcommand{\WcEttmTwoPatchtstUnder}{+28.7}
\newcommand{\WcEttmTwoPatchtstSweet}{+8.7}
\newcommand{\WcEttmTwoPatchtstOver}{+30.6}
\newcommand{\WcEttmTwoPatchtstSweetStep}{2,000}
\newcommand{\WcEttmTwoPatchtstUnderStd}{14.6}
\newcommand{\WcEttmTwoPatchtstSweetStd}{8.3}
\newcommand{\WcEttmTwoPatchtstOverStd}{3.3}

\newcommand{\WcAppliancesDlinearUnder}{+33.4}
\newcommand{\WcAppliancesDlinearSweet}{+7.4}
\newcommand{\WcAppliancesDlinearOver}{+8.4}
\newcommand{\WcAppliancesDlinearSweetStep}{4,000}
\newcommand{\WcAppliancesDlinearUnderStd}{1.0}
\newcommand{\WcAppliancesDlinearSweetStd}{2.1}
\newcommand{\WcAppliancesDlinearOverStd}{1.4}

\newcommand{\WcAppliancesPatchtstUnder}{+50.3}
\newcommand{\WcAppliancesPatchtstSweet}{+49.7}
\newcommand{\WcAppliancesPatchtstOver}{+63.0}
\newcommand{\WcAppliancesPatchtstSweetStep}{1,000}
\newcommand{\WcAppliancesPatchtstUnderStd}{4.1}
\newcommand{\WcAppliancesPatchtstSweetStd}{2.1}
\newcommand{\WcAppliancesPatchtstOverStd}{5.0}

\newcommand{\WcBdgTwoDlinearUnder}{+17.6}
\newcommand{\WcBdgTwoDlinearSweet}{+4.9}
\newcommand{\WcBdgTwoDlinearOver}{+6.0}
\newcommand{\WcBdgTwoDlinearSweetStep}{1,000}
\newcommand{\WcBdgTwoDlinearUnderStd}{0.5}
\newcommand{\WcBdgTwoDlinearSweetStd}{0.3}
\newcommand{\WcBdgTwoDlinearOverStd}{1.1}

\newcommand{\WcBdgTwoPatchtstUnder}{+20.8}
\newcommand{\WcBdgTwoPatchtstSweet}{+11.6}
\newcommand{\WcBdgTwoPatchtstOver}{+24.1}
\newcommand{\WcBdgTwoPatchtstSweetStep}{1,000}
\newcommand{\WcBdgTwoPatchtstUnderStd}{4.0}
\newcommand{\WcBdgTwoPatchtstSweetStd}{3.3}
\newcommand{\WcBdgTwoPatchtstOverStd}{2.0}

\newcommand{\WcSettings}{6}
\newcommand{\WcUnderInflatedCount}{6}
\newcommand{\WcOverInflatedCount}{6}
\newcommand{\WcUnderInflMinPt}{+0.6}
\newcommand{\WcUnderInflMaxPt}{+26.0}
\newcommand{\WcOverInflMinPt}{+1.0}
\newcommand{\WcOverInflMaxPt}{+21.9}
\newcommand{\WcUnderInflMinExclApplPatchtstPt}{+9.2}
\newcommand{\WcUnderInflApplPatchtstPt}{+0.6}
\newcommand{\WcOverInflDlinearMinPt}{+1.0}
\newcommand{\WcOverInflDlinearMaxPt}{+4.8}
\newcommand{\WcOverInflPatchtstMinPt}{+12.4}
\newcommand{\WcOverInflPatchtstMaxPt}{+21.9}

\newcommand{\WcPracticalLo}{1,000}
\newcommand{\WcPracticalHi}{20,000}
\newcommand{\WcPracticalSpreadMinPt}{3.0}
\newcommand{\WcPracticalSpreadMaxPt}{18.8}

\newcommand{\ScBdgTwoSgdmMs}{3.4}
\newcommand{\ScBdgTwoAdamMs}{3.7}

\newcommand{\ScBdgTwoRatAllSgdmMs}{3.5}
\newcommand{\ScBdgTwoRatAllAdamMs}{3.8}

\newcommand{\LtEttmTwoNWindows}{1,451}

\newcommand{\LtEttmTwoStaticQFour}{0.09}

\newcommand{\LtEttmTwoAdamHiQOne}{0.32}
\newcommand{\LtEttmTwoAdamHiQFour}{0.78}

\newcommand{\LkEtthTwoPatchtstLeaky}{+24.2}

\newcommand{\LkEtthTwoPatchtstDelayed}{\ensuremath{-}15.6}

\newcommand{\LkLeakMinPt}{+11.3}
\newcommand{\LkLeakMaxPt}{+39.8}
\newcommand{\LkEvalsetMinPt}{\ensuremath{-}23.4}
\newcommand{\LkEvalsetMaxPt}{\ensuremath{-}7.9}

\newcommand{\VpEttmTwoDlinearOracleStep}{20,000}
\newcommand{\VpEttmTwoDlinearValStep}{4,000}

\newcommand{\VpEttmTwoDlinearDelta}{+3.4}

\newcommand{\VpBdgTwoDlinearOracleStep}{1,000}
\newcommand{\VpBdgTwoDlinearValStep}{50,000}

\newcommand{\VpBdgTwoDlinearDelta}{+1.1}

\newcommand{\VpDeltaMinPt}{+0.0}
\newcommand{\VpDeltaMaxPt}{+3.4}

\newcommand{\ApplChannels}{26}
\newcommand{\ApplEnergyChannels}{2}
\newcommand{\ApplSensorChannels}{24}

\newcommand{\ApplSensorSharePct}{98.4}

\newcommand{\DataBdgTwoRepeatPct}{0.6}
\newcommand{\DataBdgTwoRatWorstRepeatPct}{44.5}

\newcommand{\pmstd}[1]{{\,\scriptsize$\pm$#1}}   

\begin{document}

\title{How Evaluation Choices Shape the Measured Benefit of Online Adaptation in
Time-Series Forecasting: Warmup, Learning-Rate Selection, and Adaptation-State Memory
}

\author{\IEEEauthorblockN{Takumi Fujimoto}
\IEEEauthorblockA{%
\textit{School of Science for Open and Environmental Systems,}\\
\textit{Graduate School of Science and Technology,}\\
\textit{Keio University}\\
Yokohama, Japan\\
takmin@keio.jp}
\and
\IEEEauthorblockN{Hiroaki Nishi}
\IEEEauthorblockA{%
\textit{School of Informatics, Management, and Human Sciences,}\\
\textit{Graduate School of Science and Technology,}\\
\textit{Keio University}\\
Yokohama, Japan\\
west@keio.jp}}

\maketitle

\begin{abstract}
Online adaptation can help edge time-series forecasting under distribution drift, but its
measured benefit is sensitive to evaluation choices. We study six public multivariate streams,
including building-sensor and smart-meter data, under a leakage-free streaming protocol. We
identify two sources of comparison bias. First, the warmup budget of the static
baseline has a two-sided effect on seed-mean curves: insufficient warmup undertrains the baseline, whereas
excessive warmup can degrade its pre-drift generalization. Across six dataset--backbone
settings, the estimated adaptation benefit changes by \WcPracticalSpreadMinPt{} to
\WcPracticalSpreadMaxPt{} percentage points (pp) over the \WcPracticalLo--\WcPracticalHi{}-step
warmup range. Second, comparing SGD with momentum (SGD$+$m) and Adam at a shared default
learning rate conflates optimizer quality with rate sensitivity. We select both the warmup
budget and each optimizer's online rate using a held-out pre-drift validation slice without
accessing test data. Under this procedure, Adam outperforms SGD$+$m in
\LrSelAdamWinsAll{} of \LrCells{} cells, and only \LrSelAdamNegCellsAll{} of its
\LrCells{} cells fall below the static baseline. Third, we characterize accuracy against adaptation-state
memory and A100-measured per-update latency for full, head-only, and calibration-based
adaptation. In the evaluated PatchTST frontier settings, several parameter-efficient variants
are nondominated on the adaptation-state-memory axis. These findings support a validation-only
commissioning procedure. Code, data, and all reported numbers: \url{https://github.com/keiotakmin/tsf-edge-adaptation}.
\end{abstract}


\section{Introduction}\label{sec:intro}
Time-series forecasting (TSF) models can be deployed on edge devices: a smart building can
predict its next-hour load from an embedded meter, and an IoT node can forecast a sensor stream
without a round trip to the cloud. Such deployments often face two coupled conditions. First,
the data distribution \emph{drifts}: occupancy patterns, seasons, retrofits, and sensor aging
can change the signal after the model is trained. This motivates \emph{on-device adaptation},
that is, continuing to update the model from the incoming stream. Second, the device is
\emph{resource-constrained}: optimizer state, peak memory, and per-update compute are
first-order deployment costs. A forecaster that adapts effectively but requires substantial
adaptation state may exceed the memory available for adaptation. The question is therefore not
only ``does adaptation help?'' but ``\emph{when} does adaptation pay, and at what resource
cost?''

Answering this question requires an evaluation protocol whose sources of bias are explicit.
The first is already known: evaluating a streaming adapter with \emph{overlapping} prediction
windows lets the model see, in a gradient step, targets that will later be scored---an
information leak formalized by DSOF~\cite{dsof2025} and by concurrent work on delayed ground
truth~\cite{actnow2024}. We use a leakage-free protocol (non-overlapping windows at stride
equal to the horizon, with each target scored \emph{before} it can enter any gradient) and
verify that our implementation reproduces the known leakage effect against a delayed-adaptation
control.

Our main contribution concerns the static comparison baseline. The reported benefit of
adaptation is measured \emph{against a baseline}---the same model without online updates---and
that baseline's quality depends on its warmup budget. On our seed-mean curves, this dependence is
\textbf{two-sided and non-monotone}. An \emph{under-warmed} baseline underfits, so adaptation
can receive credit for completing base training rather than tracking drift. An
\emph{over-warmed} baseline can generalize worse from the \emph{pre-drift} warmup segment to
the drifted test window, increasing the estimated benefit again. Within the sweep, the
estimated benefit is least sensitive to the baseline training budget near the budget that
minimizes static test error; we use that point only as a test-selected oracle reference. Its
one-sided form is related to the familiar weak-baseline critique and to observations in
test-time adaptation~\cite{zhao2023pitfalls}, but the non-monotone pattern motivates an
explicit selection rule. We therefore adapt the validation-based comparison principle of
offline optimizer benchmarking~\cite{schmidt2021descending}: early-stop warmup on a held-out
slice of the most recent \emph{pre-drift} data without accessing test data.

A second evaluation sensitivity emerged from our analysis. Comparing SGD with momentum
(SGD$+$m) and Adam at the shared rate $10^{-3}$ can make SGD$+$m appear more robust than Adam.
This is primarily a statement about \emph{learning-rate robustness}. In our grid, the two
optimizers exhibit different empirical rate ranges in which no cell falls below the static
baseline, and the shared default lies above both, further above Adam's. We select each
optimizer's rate by \emph{rehearsing} online adaptation on the same held-out pre-drift slice.
This selection substantially changes the observed cell-wise comparison.

Under this validation-only protocol, we characterize accuracy against adaptation-state memory
and A100-measured update latency for several adaptation strategies. All tuned choices are made
per device at commissioning from that device's own pre-drift data, with no test data and no
cross-device coordination. We additionally evaluate smart-meter subsets, a 280-meter site, and
a 240-meter fleet spanning 18 of 19 BDG2 sites (\S\ref{sec:frontier}).

\smallskip\noindent\textbf{Contributions.}
\begin{itemize}
\item \textbf{C1} --- The baseline's \textbf{warmup budget} is an evaluation sensitivity
distinct from target leakage and is two-sided and non-monotone on the seed-mean curves of our
sweep, most clearly for PatchTST. Validation
early-stopping on held-out pre-drift data selects a reference whose estimated benefit differs
from the test-selected oracle by at most \VpDeltaMaxPt\,pp in the six examined panels
(\S\ref{sec:c1}).
\item \textbf{C2} --- A shared \textbf{online learning-rate default} can bias the SGD$+$m
versus Adam comparison. Optimizer-specific rate rehearsal substantially reduces this
default-rate asymmetry and changes the observed winner counts (\S\ref{sec:lrfair}).
\item \textbf{C3} --- A validation-selected \textbf{accuracy--adaptation-state-memory--compute}
characterization over the evaluated parameter subsets and optimizers. In the PatchTST frontier
settings, several parameter-efficient variants are nondominated on the adaptation-state-memory
axis (\S\ref{sec:frontier}); \S\ref{sec:recipe} distils conditional guidance from these
measurements.
\end{itemize}

\section{Related Work}
\noindent\textbf{Evaluation leakage in online TSF.}
DSOF~\cite{dsof2025} and concurrent work on delayed-ground-truth protocols~\cite{actnow2024}
identify that overlapping evaluation windows let a streaming model train on targets it will
later be scored on, inflating the apparent benefit of adaptation; recent test-time adapters
(\mbox{$\delta$-Adapter}~\cite{deltaadapter2026}, SOLID~\cite{solid2024},
TAFAS~\cite{tafas2025}) operate in related streaming settings, and further concurrent work
adapts only on \emph{matured} ground truth~\cite{principledtta2026}---the same principle as
the delayed control arm we use in \S\ref{sec:setup}. We adopt this line's protocol rather than
extend it. The separate evaluation sensitivity we study is the baseline's warmup budget
(\S\ref{sec:intro}, \S\ref{sec:c1}), which, to our knowledge, prior work in this line has
not isolated.

\smallskip\noindent\textbf{Parameter-efficient and test-time adaptation.}
PETSA~\cite{petsa2025} proposes parameter-efficient input/output calibration for online
forecasting; our \texttt{calib} strategy is a simplified calibration variant inspired by
PETSA---a per-channel affine input calibration ($2C$ parameters) plus the linear output head,
in place of PETSA's low-rank additive modules and dedicated calibration loss. We use it as one
resource-constrained adaptation point and do not compare it with the official implementation
(\S\ref{sec:discussion}). Broader online/continual adaptation for TSF
(FSNet~\cite{pham2023fsnet}, OneNet~\cite{zhang2023onenet}, and the detect-then-adapt /
proactive-drift lines D$^3$A~\cite{d3a2024} and PROCEED~\cite{proceed2025}) provides context
for our drift-triggered schedule; we vary adaptation strategy \emph{within} fixed backbones
rather than benchmark against these methods. OneNet, for example, maintains two model
instances, a cost outside our adaptation-state accounting.

\smallskip\noindent\textbf{Building-energy forecasting at the edge.}
Online adaptation for building-load forecasting exists as an algorithmic line---online
adaptive RNNs under concept drift~\cite{fekri2021online} and subsequent online
ensembles---but without reporting the adaptation-state memory and per-update compute metrics
used here. Symmetrically, \emph{training on} real smart meters has recently been shown
feasible: federated split learning trains load forecasters on meters with 192\,kB of
SRAM~\cite{li2024edgemeter}, and one-off on-device training of PV forecasters has been
demonstrated on a commercial meter~\cite{huang2025ondevice}. Neither, however, adapts online
under drift. The broader on-device/TinyML continual-learning literature is largely
vision-centric. We did not identify work jointly evaluating online TSF adaptation on edge
hardware under drift with the memory and compute criteria used here. This gap
motivates our study.

\smallskip\noindent\textbf{When to adapt.}
``Adapt Only When It Pays''~\cite{adowip2026} frames adaptation as a budgeted decision, with a
regret guarantee, over \emph{whether} to spend an update. Our work is complementary: given
that one adapts, we compare optimizer/rate and parameter-subset choices and their
adaptation-state memory cost. The regimes where we find small benefits further qualify that
\emph{whether} question on independent data.

\smallskip\noindent\textbf{Fair-evaluation methodology.}
The fair-benchmarking tradition~\cite{schmidt2021descending} stresses tuned baselines and
multiple seeds for \emph{offline} optimizer comparison; our warmup and learning-rate protocols
are motivated by the same validation-based comparison principle. The closest prior warning is
on the test-time-adaptation side: TTAB~\cite{zhao2023pitfalls} finds TTA benefit strongly
dependent on the quality of the model being adapted, and identifies hyperparameter and model
selection as difficult parts of a fair protocol. That dependence is \emph{monotone} in source
quality; the two-sided form we find (\S\ref{sec:intro}, \S\ref{sec:c1}) makes a generic
recommendation to ``train the baseline well'' insufficient, and motivates the
forecasting-specific selection procedure studied here.

\section{Experimental Setup}\label{sec:setup}
\noindent\textbf{Data.}
Six public multivariate series, selected to include differing observed drift patterns, two of
them building data (a mixed sensor stream and a pure energy-meter corpus): the four ETT subsets
(ETTh1, ETTh2, ETTm1, ETTm2; electricity-transformer temperature/load)~\cite{zhou2021informer},
the UCI \emph{Appliances} dataset, a multivariate building-sensor stream comprising appliance
and lighting load together with indoor/outdoor temperature, humidity, and weather variables
~\cite{candanedo2017data}\footnote{Appliances is a \ApplChannels-channel
building-\emph{sensor} stream: \ApplEnergyChannels{} energy channels (appliance and lighting
load) and \ApplSensorChannels{} indoor/outdoor temperature, humidity and weather channels (the
two random columns UCI supplies for feature-selection tests are dropped). All channels are
forecast, so \ApplSensorSharePct\% of the $z$-normalized test variance over which the MSE is
computed comes from sensor rather than energy channels, and the drift is dominated by a
winter-to-spring temperature ramp.
BDG2 is the pure energy-meter corpus.}, and a subset of the \emph{Building Data Genome~2}
(BDG2) corpus~\cite{miller2020building}. The BDG2 subset is site ``Rat,'' the corpus's largest
by meter count; from it we take the 15 buildings with the least missingness and two years of
hourly meter readings. Missing BDG2 readings are imputed per meter by forward filling. If a
leading missing prefix remains before a meter's first valid observation, it is backward-filled
from that first observation; thus, after the first valid observation, no imputed value uses a
later time point. In the main Rat least-missing subset, backward filling is a no-op.
(\S\ref{sec:frontier} additionally evaluates two further sites, an \emph{anti-selection} subset
of the most-missing meters, the full 280-meter Rat site, and a 240-meter fleet.) Each series is
split by time into a pre-drift warmup pool (first half) and a streamed test region (second half),
and $z$-normalized using statistics computed on the warmup pool only. All Appliances results
reported below are multivariate MSEs over the \ApplChannels{} channels; we therefore use this
dataset as a mixed building-sensor setting rather than as an appliance-load-only evaluation.

\smallskip\noindent\textbf{Backbones.}
We use two forecasting backbones with contrasting parameterization:
\textbf{DLinear}~\cite{zeng2023transformers} (a decomposition-linear model with a
trend/seasonal moving-average split and two linear maps), and a compact,
channel-independent \textbf{PatchTST}~\cite{nie2023patchtst} (patch embedding, a 2-layer
transformer encoder, and a linear head, plus a per-channel input affine calibration targeted by
our PEFT strategy). DLinear provides a linear baseline; PatchTST provides a compact transformer
baseline.

\smallskip\noindent\textbf{Quality metric and grid.}
Both errors we compare are online (prequential) MSEs over the streamed test region, and every
benefit figure in this paper is the \textbf{adaptation benefit}
$(\mathrm{MSE}_{\mathrm{static}}-\mathrm{MSE}_{\mathrm{adapt}})/\mathrm{MSE}_{\mathrm{static}}$
in \%, so a positive value means adaptation beats the static baseline and a larger one means a
larger relative improvement. A difference between two such percentages is reported in percentage
points (pp). A \textbf{configuration} is one (dataset, backbone, horizon $H$, lookback $L$)
tuple, and a \textbf{cell} is one seed-specific realization of that configuration. The optimizer
grid is 6~datasets $\times$ 2~backbones $\times$ $H\in\{24,48,96\}$ $\times$
$L\in\{96,192\}$, i.e., \LrConfigs{} configurations at five seeds each, for \LrCells{} cells.
Selection and aggregation use different units: Table~\ref{tab:confound} selects its oracle
reference on the mean curve across three seeds; Table~\ref{tab:lrfair} reports cell-level counts
and seed-majority configuration counts; and Table~\ref{tab:bdgscale} selects the optimizer
separately for each seed. Unless stated otherwise, dispersion is reported descriptively as the
standard deviation across the stated seeds, not as a confidence interval.

\smallskip\noindent\textbf{Leakage-free streaming protocol.}
After warmup, we stream the test region with \textbf{non-overlapping} windows at stride equal
to the horizon $H$. At each step we predict the next $H$ steps from the previous $L$-step
lookback in one forward pass (both backbones emit all $H$ steps directly, not autoregressively),
\textbf{score the prediction against the revealed truth, and only then adapt} on that
(lookback, truth) pair. Because every scored target is evaluated before it can appear in any
gradient, the protocol is leakage-free by construction.

\smallskip\noindent\textbf{Verifying the protocol.}
We reproduce leakage only to verify our implementation. To isolate it, we add a \emph{delayed}
control arm: stride-1 evaluation in which the model adapts at every step only on the most
recent \emph{fully-revealed} window, so no future evaluation target can enter any gradient.
Against that control, target leakage increases the reported benefit by \LkLeakMinPt{} to
\LkLeakMaxPt{}\,pp, reversing its sign on ETTh2/PatchTST
(\LkEtthTwoPatchtstLeaky\% leaky vs.\ \LkEtthTwoPatchtstDelayed\% delayed). The residual
stride-1-vs-stride-$H$ difference (\LkEvalsetMinPt{} to \LkEvalsetMaxPt{}\,pp over this
control's four settings, ETTh2 and Appliances $\times$ both backbones) is consistent with the
combined effect of denser evaluation and an $H\times$ higher update count. All later results use
the leakage-free protocol; the warmup sensitivity of \S\ref{sec:c1} is analyzed separately.

\smallskip\noindent\textbf{Adaptation strategies and schedules.}
We compare parameter-subset and optimizer combinations: \texttt{static} (no adaptation),
\texttt{full$\cdot$SGD+m}, \texttt{full$\cdot$Adam}~\cite{kingma2015adam},
\texttt{head$\cdot$SGD+m} (output head only; for DLinear, whose two linear maps have no
separate head, this updates the trend-branch map), and \texttt{calib$\cdot$SGD+m}, a simplified
PatchTST per-channel input/output calibration variant. \S\ref{sec:frontier} completes the
cross product by pairing \texttt{head} and \texttt{calib} with Adam as well. Update schedules
are every-step, periodic (every $k$), and drift-triggered (adapt when the current error exceeds
$\tau\times$ a running error EMA). We sweep, rather than tune, $k$ and $\tau$
($k\in\{1,2,4,8,16\}$, $\tau\in\{1.1,1.3,1.5,2.0,3.0\}$) to trace each schedule's
quality-vs-update-budget curve, and compare the two curves at matched update fraction.

\smallskip\noindent\textbf{Resource metrics.}
For each strategy we log \textbf{optimizer-state bytes} (measured, not assumed, where the
state is not a fixed multiple of the trainable-parameter bytes), \textbf{per-update
wall-clock}, an illustrative time--power calculation (wall-clock $\times$ a 5\,W device
assumption), and the trainable-parameter count. Wall-clock is the median over updates after a
discarded warm-up prefix, minimized over repeats that interleave every strategy: at batch size~1
the update is launch-latency bound, so a single sequential pass can be dominated by host
contention and warm-up rather than optimizer computation. The C3 memory axis is
\emph{adaptation-state memory}: gradient buffer plus optimizer state (8\,B/param for SGD$+$m,
12 for Adam in fp32; momentum-free SGD would be 4, but we do not report it). This accounting
excludes model weights, activations, and runtime-memory overhead.

\smallskip\noindent\textbf{Validation-selected warmup.}
The warmup budget is chosen by \textbf{early-stopping on a held-out pre-drift validation slice}
(the most recent 20\% of the warmup pool), i.e., without touching the test region. Two grids of
warmup budgets appear in this paper, and every arm of a given comparison shares one. The
\emph{study} grid of \S\ref{sec:c1} runs over \WcMilestoneCount{} milestones from
\WcMilestoneLo{} to \WcMilestoneHi{} steps (Fig.~\ref{fig:warmup};
Table~\ref{tab:confound}, whose ``under'' and ``over'' columns are its two ends); it is deliberately wider than any deployment would sweep, because exhibiting a
two-sided sensitivity requires both extremes. Both selection arms read that same grid at the
same milestones: the validation criterion above, and an oracle reference (the budget minimizing
static \emph{test} error) used only to quantify the validation criterion's proximity
(\S\ref{sec:c1}). The oracle never picks above \GridWarmCapStep{} steps, and neither does the
validation early-stop in five of the six dataset$\times$backbone panels. The exception is
BDG2/DLinear, where the validation curve is still improving at the top of the grid
(\S\ref{sec:c1}). The \emph{deployment} grid, applied per cell by every downstream measurement
(frontier, schedules, and the optimizer grid), is the practical \GridWarmGridCount{} milestones
from \GridWarmGridLo{} to \GridWarmCapStep{} steps; all of them run the identical criterion, so
their baselines coincide exactly. Unlike the rate grid, it is capped rather than bracketed:
\GridWarmCapCells{} of \GridCells{} cells select the top budget, so the selected budget is
truncated by the grid for those cells.
The slice is deliberately the \emph{most recent} pre-drift data and is held out of all base
training. We use it as a proxy for the forthcoming stream, and both selection procedures
(warmup early-stopping here and rate rehearsal below) require a stream that the model has not
fitted. The cost is reduced training-data freshness (\S\ref{sec:discussion}).

\smallskip\noindent\textbf{Validation-selected online LR (rehearsal).}
The online learning rate is selected without test data in the same spirit: after warmup, we
\emph{rehearse} online adaptation over the held-out pre-drift validation slice, using the
identical score-then-adapt streaming. The rehearsal runs at each of ten candidate rates on a
$\{1,3\}{\times}10^{k}$ grid spanning $3{\times}10^{-6}$ to $10^{-1}$, separately per strategy
and per optimizer. We then deploy the rate with the lowest validation online MSE, held constant
for the whole stream (no decay or adaptive schedule). The rehearsal is a one-off commissioning cost,
incurred wherever the base model is warmed up (on-device or tethered): ten short validation
streams per optimizer, in our splits about two online passes over a stream of length comparable to the test region.
Adam's momentum constants stay at their defaults
$(\beta_1,\beta_2)=(0.9,0.999)$. All results use three seeds for the warmup studies and five for the
optimizer grid, whose full design (\LrCells{} cells) is covered by the rate sweep of
\S\ref{sec:lrfair}. The few measurements that depart from this are noted where they appear.

\section{Results}

\subsection{C1 --- Warmup Sensitivity and a Validation-Selected Protocol}\label{sec:c1}

\begin{figure*}[t]
\centering
\includegraphics[width=\textwidth]{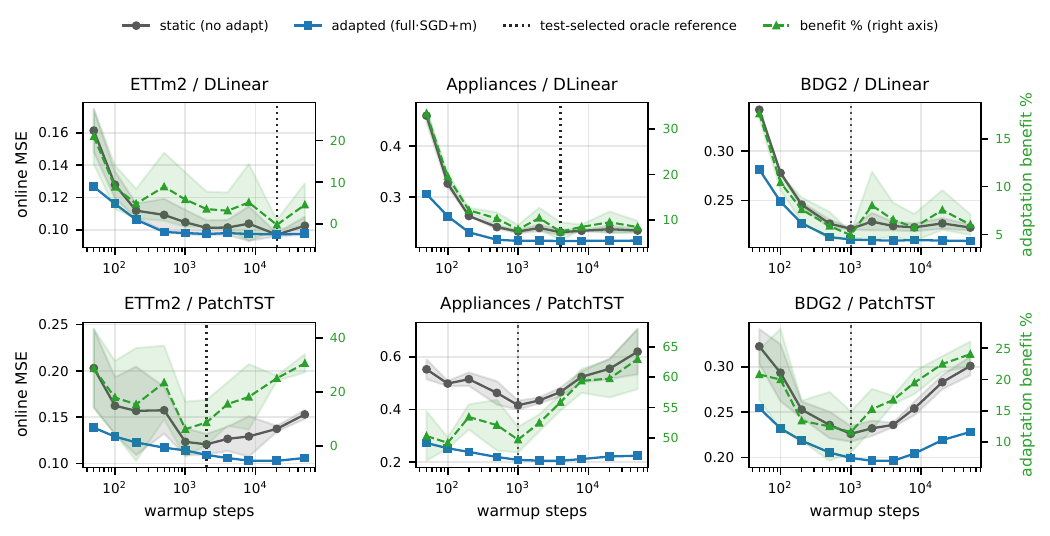}
\caption{Warmup sensitivity on the study milestone grid (three seeds; mean\,$\pm$\,standard deviation).
In each dataset--backbone panel, gray and blue curves show the online test MSE of the static and
full$\cdot$SGD$+$m adapted models, respectively (left axis), and the green curve shows adaptation
benefit (right axis). The black dotted rule marks the test-selected oracle warmup.}
\label{fig:warmup}
\end{figure*}

\noindent\textbf{Warmup sensitivity is two-sided and non-monotone on seed-mean curves.}
Sweeping the baseline warmup budget from \WcMilestoneLo{} to \WcMilestoneHi{} steps (three seeds,
six dataset$\times$backbone panels; Fig.~\ref{fig:warmup}), the static baseline's test error is
U-shaped, and the estimated adaptation benefit is larger at both ends of the sweep
(Table~\ref{tab:confound}). All readings in this paragraph are on seed-mean curves.
\textbf{Under-warming} increases the estimated benefit in \WcUnderInflatedCount{} of
\WcSettings{} panels by \WcUnderInflMinPt{} to \WcUnderInflMaxPt{}\,pp: an under-warmed baseline
underfits, and the measured benefit can include completion of base training. \textbf{Over-warming}
increases it in \WcOverInflatedCount{} of \WcSettings{} panels by \WcOverInflMinPt{} up to
\WcOverInflMaxPt{}\,pp, as the baseline overfits the pre-drift segment. In both cases the
static error rises faster than the adapted error (Fig.~\ref{fig:warmup}): online updates recover
part of the deficit of an underfit or overfit starting point, so the benefit follows the static
curve's U~shape. Whether the over-warming deficit is memorized pre-drift level and seasonality
is untested. With three seeds, the two sides are not equally well resolved. The over-warming
increase is \WcOverInflPatchtstMinPt{} to \WcOverInflPatchtstMaxPt{}\,pp in the three PatchTST
panels, above the seed standard deviations of Table~\ref{tab:confound}, but only
\WcOverInflDlinearMinPt{} to \WcOverInflDlinearMaxPt{}\,pp in the DLinear panels, comparable to
them. The under-warming increase is at least \WcUnderInflMinExclApplPatchtstPt\,pp and above
them in five panels, and \WcUnderInflApplPatchtstPt\,pp, within them, on Appliances/PatchTST.
The two-sided pattern is thus clear of seed spread only for PatchTST on ETTm2 and BDG2. We use
the warmup budget with minimum static test error only as an oracle reference. Nor does the effect occur only at the two extreme budgets.
Within our \WcPracticalLo--\WcPracticalHi{}-step practical sweep, the estimated benefit still
moves by \WcPracticalSpreadMinPt{} to \WcPracticalSpreadMaxPt{}\,pp across the six panels.
Changing only the warmup budget in this controlled sweep can therefore move the estimated
benefit by more than the gap between the optimizers compared in \S\ref{sec:lrfair}.

\begin{table}[t]
\centering
\caption{\textbf{Warmup sensitivity (C1).} Adaptation benefit (\%; mean\,$\pm$\,population standard
deviation over three seeds) at \WcMilestoneLo{} steps, at the test-selected oracle reference, and
at \WcMilestoneHi{} steps. The oracle reference is the warmup step that minimizes the
seed-averaged static-test MSE; the oracle column reports both the benefit and that step.}
\label{tab:confound}
\begin{tabular}{l r r r}
\toprule
dataset / backbone & under & \emph{oracle ref.} & over \\
\midrule
ETTm2 / DLinear & \WcEttmTwoDlinearUnder\pmstd{\WcEttmTwoDlinearUnderStd} & \emph{\WcEttmTwoDlinearSweet}\pmstd{\WcEttmTwoDlinearSweetStd} @\WcEttmTwoDlinearSweetStep & \WcEttmTwoDlinearOver\pmstd{\WcEttmTwoDlinearOverStd} \\
ETTm2 / PatchTST & \WcEttmTwoPatchtstUnder\pmstd{\WcEttmTwoPatchtstUnderStd} & \emph{\WcEttmTwoPatchtstSweet}\pmstd{\WcEttmTwoPatchtstSweetStd} @\WcEttmTwoPatchtstSweetStep & \WcEttmTwoPatchtstOver\pmstd{\WcEttmTwoPatchtstOverStd} \\
Appl.\ / DLinear & \WcAppliancesDlinearUnder\pmstd{\WcAppliancesDlinearUnderStd} & \emph{\WcAppliancesDlinearSweet}\pmstd{\WcAppliancesDlinearSweetStd} @\WcAppliancesDlinearSweetStep & \WcAppliancesDlinearOver\pmstd{\WcAppliancesDlinearOverStd} \\
Appl.\ / PatchTST & \WcAppliancesPatchtstUnder\pmstd{\WcAppliancesPatchtstUnderStd} & \emph{\WcAppliancesPatchtstSweet}\pmstd{\WcAppliancesPatchtstSweetStd} @\WcAppliancesPatchtstSweetStep & \WcAppliancesPatchtstOver\pmstd{\WcAppliancesPatchtstOverStd} \\
BDG2 / DLinear & \WcBdgTwoDlinearUnder\pmstd{\WcBdgTwoDlinearUnderStd} & \emph{\WcBdgTwoDlinearSweet}\pmstd{\WcBdgTwoDlinearSweetStd} @\WcBdgTwoDlinearSweetStep & \WcBdgTwoDlinearOver\pmstd{\WcBdgTwoDlinearOverStd} \\
BDG2 / PatchTST & \WcBdgTwoPatchtstUnder\pmstd{\WcBdgTwoPatchtstUnderStd} & \emph{\WcBdgTwoPatchtstSweet}\pmstd{\WcBdgTwoPatchtstSweetStd} @\WcBdgTwoPatchtstSweetStep & \WcBdgTwoPatchtstOver\pmstd{\WcBdgTwoPatchtstOverStd} \\
\bottomrule
\end{tabular}
\end{table}

\begin{figure*}[t]
\centering
\includegraphics[width=\textwidth]{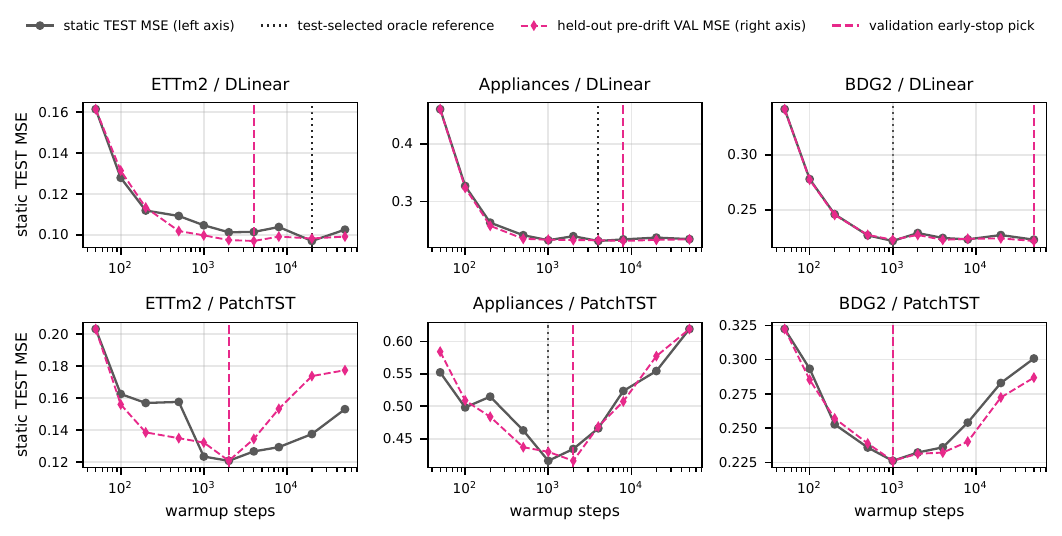}
\caption{Validation-only warmup selection on the study milestone grid and panel layout of
Fig.~\ref{fig:warmup} (three-seed means). Gray and magenta curves show
static test MSE (left axis) and held-out pre-drift validation MSE (right axis), respectively. The
magenta dashed and black dotted rules mark the validation early-stop and test-selected oracle
warmups. The validation-MSE scale is suppressed because only its argmin is used for selection.}
\label{fig:valprotocol}
\end{figure*}

\smallskip\noindent\textbf{A validation-only selection procedure.}
The oracle reference above uses test data. Early-stopping warmup on the held-out \emph{pre-drift}
validation slice changes the estimated benefit by \VpDeltaMinPt{} to \VpDeltaMaxPt{}\,pp relative
to that reference across all \WcSettings{} panels (three seeds; both selections read on
seed-mean curves; Fig.~\ref{fig:valprotocol}). This residual is consistent with distributional
change between the validation slice and the drifted test, so C1 provides a validation-only
selection procedure rather than merely a diagnostic. The criterion does \emph{not} reproduce
the budget itself. On BDG2/DLinear it picks \VpBdgTwoDlinearValStep{} steps against an oracle
reference at \VpBdgTwoDlinearOracleStep{}, five milestones apart, yet changes the estimated
benefit by only \VpBdgTwoDlinearDelta\,pp; the largest residual,
\VpEttmTwoDlinearDelta\,pp on ETTm2/DLinear at \VpEttmTwoDlinearValStep{} against
\VpEttmTwoDlinearOracleStep{}, is similar. In both cases the static-test curve is flat enough
that a distant budget has little effect.

\subsection{C2 --- A Shared Learning-Rate Default Can Bias Optimizer Comparisons}\label{sec:lrfair}

\begin{figure*}[t]
\centering
\includegraphics[width=\textwidth]{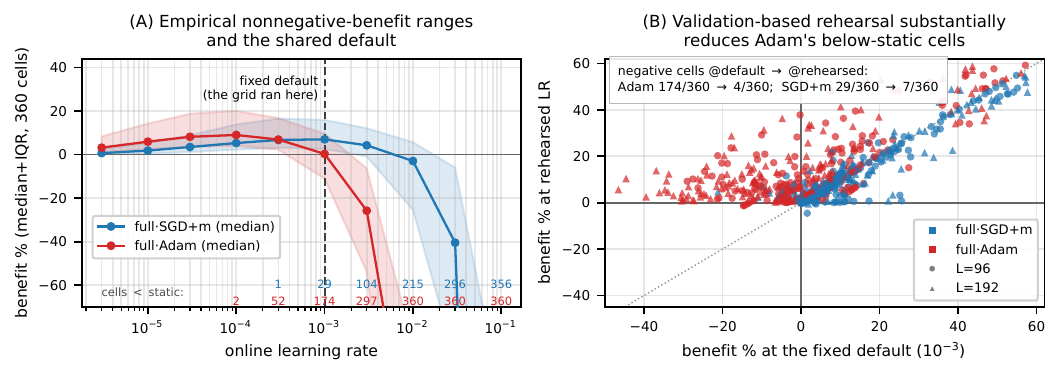}
\caption{Learning-rate sensitivity across all \LrCells{} cells. \textbf{(A)} Adaptation benefit by
online learning rate, summarized by the median and interquartile range across cells; numbers along
the bottom give the count of cells below the static baseline at each rate. The dashed line marks
the shared default, $10^{-3}$. \textbf{(B)} Per-cell benefit at the shared default ($x$-axis)
versus the validation-rehearsed rate ($y$-axis); color denotes optimizer and marker shape denotes
lookback length.}
\label{fig:regime}
\end{figure*}

\noindent\textbf{The apparent asymmetry at the default rate.}
Over all \GridCells{} cells, with \emph{both} online optimizers at the commonly used rate of
$10^{-3}$,\footnote{Adam's original default~\cite{kingma2015adam}, the default of both
\texttt{torch.optim.Adam} and \texttt{torch.optim.SGD} (v2.7), and OneNet's~\cite{zhang2023onenet}
test-time rate; FSNet~\cite{pham2023fsnet} defaults to $10^{-4}$, which yields the
opposite conclusion (Fig.~\ref{fig:regime}A). A shared fixed rate can bias the comparison.} the
observed cell counts are strongly asymmetric, and the asymmetry concerns \emph{robustness}:
\texttt{full$\cdot$SGD+m} is worse than the static baseline in
\GridSgdmNegCells/\GridCells{} cells and reaches a minimum of \GridSgdmFloor\%, whereas
\texttt{full$\cdot$Adam} is worse in
\GridAdamNegCells/\GridCells{} (\GridAdamNegPct\%) and reaches a minimum of
\GridAdamWorst\%---for one extra state copy per parameter.
\GridUnanimousConfigs{} of \GridConfigs{} configurations agree on the winner across five
seeds, and the asymmetry sharpens with lookback (Adam negative in
\GridLOneNineTwoAdamNegPct\% of $L{=}192$ cells vs.\ \GridLNineSixAdamNegPct\% at $L{=}96$).
Read as a property of the optimizers, this grid would favor SGD$+$m under the shared-rate
evaluation. The rest of this section shows that the pattern is primarily a statement about one
rate.

\smallskip\noindent\textbf{The asymmetry depends on the rate.}
For \emph{every} cell we swept a ten-point rate grid ($3{\times}10^{-6}$--$10^{-1}$) per
optimizer, logging validation-rehearsal and test performance at every rate
(Fig.~\ref{fig:regime}A); read at $10^{-3}$, the sweep reproduces the default-rate grid of the
preceding paragraph cell-for-cell. In this grid, both optimizers exhibit empirical
nonnegative-benefit ranges, but the ranges are offset by about a grid step: SGD$+$m has no
below-static cells from $3{\times}10^{-6}$ through $10^{-4}$ and Adam only through
$3{\times}10^{-5}$, after which Adam degrades more quickly---already
\LrAdamAtOneEMinusThreeNegCells{} negative cells at $10^{-3}$ against SGD$+$m's
\LrSgdmAtOneEMinusThreeNegCells{}, and every one of the \LrCells{} cells from $10^{-2}$ up
(per-rate counts along the bottom of Fig.~\ref{fig:regime}A). The default $10^{-3}$ thus lies
above both ranges, but one grid step further above Adam's, where degradation is already
widespread. Even a
fixed $10^{-4}$, with no per-cell tuning, changes the observed cell-wise comparison: Adam wins
\LrAdamWinsAtOneEMinusFour{} of \LrCells{} cells (versus
\LrAdamWinsAtOneEMinusThree{} at $10^{-3}$), with \LrAdamAtOneEMinusFourNegCells{} negative
cells and a mean benefit of \LrAdamAtOneEMinusFourMean\%. This is the highest mean benefit
among the fixed-rate settings evaluated here; SGD$+$m's best is
\LrSgdmAtThreeEMinusFourMean\% at $3{\times}10^{-4}$. The error trajectory at large rates is consistent with a loss of tracking stability: on the
longest stream (ETTm2, \LtEttmTwoNWindows{} windows; seed 0), Adam@$10^{-2}$'s per-window error
rises from \LtEttmTwoAdamHiQOne{} in the first quarter to \LtEttmTwoAdamHiQFour{} in the last,
against \LtEttmTwoStaticQFour{} for the static model.

\smallskip\noindent\textbf{The comparison at rehearsed rates.}
Table~\ref{tab:lrfair} reads the same cells at three per-cell rate choices. At the
\emph{rehearsed} rate (the validation-based procedure of \S\ref{sec:setup}), Adam wins
\LrSelAdamWinsAll{} of \LrCells{} cells overall. At the three evaluated horizons, it wins
\LrHTwoFourSelAdamWins, \LrHFourEightSelAdamWins{} and \LrHNineSixSelAdamWins{} of
\LrHTwoFourCells{} cells at $H{=}24$, $48$, and $96$, respectively. Cells of one
configuration share data, so we also compare seed-averaged configurations: Adam's gap over
SGD$+$m is positive in \LrSelCfgMeanGapPosCfgs{} of \LrConfigs{}, with a median of
\LrSelCfgMeanGapMedianPt\,pp (order-statistic 95\% interval \LrSelCfgMeanGapCiLoPt{} to
\LrSelCfgMeanGapCiHiPt\,pp; two-sided Wilcoxon signed-rank $p{=}\LrSelCfgMeanGapWilcoxonP$), and in
\LrSelDsBbMeanGapPos{} of \LrSelDsBbTotal{} dataset--backbone pairs. Configurations still share
datasets, so we read this test descriptively. Adam's below-static count
falls from \LrAdamAtOneEMinusThreeNegCells{} at the default to \LrSelAdamNegCellsAll{} of
\LrCells{} cells at the rehearsed rate; the latter are all ETTm2/DLinear, with a worst value of
\LrSelAdamMinAll\% (Fig.~\ref{fig:regime}B). In \LrSelAdamVsOrcSgdmWins{} of \LrCells{} cells,
rehearsed Adam even beats SGD$+$m at its test-selected oracle rate, an unavailable reference at
deployment (median \LrSelAdamVsOrcSgdmMedianPt\,pp).
Adam's rehearsed rate is at most $3{\times}10^{-4}$ in
\LrAdamSelLeqThreeEMinusFourCells{} of \LrCells{} cells ($3{\times}10^{-5}$ or $10^{-4}$ in
\LrAdamSelTenthToThirtiethCells) and shifts one notch lower at
$L{=}192$, where the adapted parameter count roughly doubles; symmetrically, the
\LrAdamSelGeqOneEMinusThreeCells{} cells where it does pick $10^{-3}$ or higher all sit at
$H\in\{48,96\}$, where the stream contains $2$--$4\times$ fewer updates. Both shifts are
consistent with dependence on the cumulative amount of adaptation exposure. This also explains
the lookback effect: at the default, longer lookbacks move Adam further outside its empirical
nonnegative-benefit range; at rehearsed rates they narrow Adam's margin (median
\LrLNineSixSelMedianGapPt{} to \LrLOneNineTwoSelMedianGapPt\,pp; \LrSelMedianGapPtAll\,pp over
all cells) without changing its sign.

\begin{table}[t]
\centering
\caption{\textbf{Learning-rate sensitivity (C2) over the same cells, evaluated at three per-cell rate
choices.} Values are adaptation benefit (\%) relative to the static baseline ($>$0 is better);
negative cells are reported as count (worst value). The last column is the median per-cell
Adam$-$SGD$+$m difference, and ``configs'' counts seed-majority winners. ``Default'' fixes both
optimizers at $10^{-3}$; ``rehearsed'' selects the rate using held-out pre-drift validation;
``oracle'' selects the best rate on the test stream.}
\label{tab:lrfair}
\footnotesize\setlength{\tabcolsep}{3.5pt}
\begin{tabular}{l cc c c r}
\toprule
& \multicolumn{2}{c}{negative cells (worst \%)} & wins & configs & median \\
\cmidrule(lr){2-3}
rate choice & SGD$+$m & Adam & S/A & S/A & A$-$S (pp) \\
\midrule
\multicolumn{6}{l}{\emph{$L{=}96$ (36 configurations $\times$ 5 seeds $=$ 180 cells)}} \\
default $10^{-3}$ & \LrLNineSixFixedSgdmNegCells{} (\LrLNineSixFixedSgdmMin) & \LrLNineSixFixedAdamNegCells{} (\LrLNineSixFixedAdamMin) & \LrLNineSixFixedSgdmWins/\LrLNineSixFixedAdamWins & \LrLNineSixFixedCfgSgdmWins/\LrLNineSixFixedCfgAdamWins & \LrLNineSixFixedMedianGapPt \\
rehearsed & \LrLNineSixSelSgdmNegCells{} (\LrLNineSixSelSgdmMin) & \LrLNineSixSelAdamNegCells{} (\LrLNineSixSelAdamMin) & \LrLNineSixSelSgdmWins/\LrLNineSixSelAdamWins & \LrLNineSixSelCfgSgdmWins/\LrLNineSixSelCfgAdamWins & \LrLNineSixSelMedianGapPt \\
oracle & \LrLNineSixOrcSgdmNegCells{} (\LrLNineSixOrcSgdmMin) & \LrLNineSixOrcAdamNegCells{} (\LrLNineSixOrcAdamMin) & \LrLNineSixOrcSgdmWins/\LrLNineSixOrcAdamWins & \LrLNineSixOrcCfgSgdmWins/\LrLNineSixOrcCfgAdamWins & \LrLNineSixOrcMedianGapPt \\
\midrule
\multicolumn{6}{l}{\emph{$L{=}192$ (36 configurations $\times$ 5 seeds $=$ 180 cells)}} \\
default $10^{-3}$ & \LrLOneNineTwoFixedSgdmNegCells{} (\LrLOneNineTwoFixedSgdmMin) & \LrLOneNineTwoFixedAdamNegCells{} (\LrLOneNineTwoFixedAdamMin) & \LrLOneNineTwoFixedSgdmWins/\LrLOneNineTwoFixedAdamWins & \LrLOneNineTwoFixedCfgSgdmWins/\LrLOneNineTwoFixedCfgAdamWins & \LrLOneNineTwoFixedMedianGapPt \\
rehearsed & \LrLOneNineTwoSelSgdmNegCells{} (\LrLOneNineTwoSelSgdmMin) & \LrLOneNineTwoSelAdamNegCells{} (\LrLOneNineTwoSelAdamMin) & \LrLOneNineTwoSelSgdmWins/\LrLOneNineTwoSelAdamWins & \LrLOneNineTwoSelCfgSgdmWins/\LrLOneNineTwoSelCfgAdamWins & \LrLOneNineTwoSelMedianGapPt \\
oracle & \LrLOneNineTwoOrcSgdmNegCells{} (\LrLOneNineTwoOrcSgdmMin) & \LrLOneNineTwoOrcAdamNegCells{} (\LrLOneNineTwoOrcAdamMin) & \LrLOneNineTwoOrcSgdmWins/\LrLOneNineTwoOrcAdamWins & \LrLOneNineTwoOrcCfgSgdmWins/\LrLOneNineTwoOrcCfgAdamWins & \LrLOneNineTwoOrcMedianGapPt \\
\bottomrule
\end{tabular}
\end{table}

\smallskip\noindent\textbf{Limits of the rehearsal criterion.}
What SGD$+$m retains is robustness, not quality, and one state copy per parameter
instead of two. Rehearsal trades a little of that robustness for quality. On the calmer pre-drift slice a
rate beyond that range can \emph{win the rehearsal} and still destabilize on the drifted stream,
so SGD$+$m's rehearsal reaches into the top two rates of the grid ($3{\times}10^{-2}$ and
$10^{-1}$) in \LrSgdmSelExtCells{} of
\LrCells{} cells and \LrSgdmSelNegCellsAll{} of its picks land below static (worst
\LrSgdmSelMinAll\%, \LrSgdmSelNegAppliancesDlinearCells{} of them on the most drift-heavy
pairing, Appliances/DLinear), against \LrSelAdamNegCellsAll{} cells (worst
\LrSelAdamMinAll\%) for Adam (Fig.~\ref{fig:regime}B). Momentum removes the severe form of
this failure: with plain SGD the same protocol produced diverged streams, whereas
none of the SGD$+$m picks diverges and the worst case is a few
pp rather than a collapse. The pre-drift slice can select strong validation quality within an
empirical nonnegative-benefit range, but it does not certify stability under drift. A
conservative variant---the smallest rate within 2\% of the best validation MSE---eliminates
below-static picks in this grid, but it alters \LrTolGuardChangedCells{} of \LrCells{} picks at
a median cost of \LrTolGuardMedianCostPt\,pp, so we report the unmodified rule. The results
suggest a division of roles: SGD$+$m is more tolerant of a mis-set rate, whereas Adam achieves
higher benefits after rate rehearsal in this grid.

\smallskip\noindent\textbf{Regime probes do not predict the optimizer.}
Three cheap probes on the frozen base model---stream noise, gradient consistency, and a
post-hoc drift ratio---yield no usable optimizer guidance: at the rehearsed rate the choice
they were meant to inform has largely dissolved, and at the fixed default they separate the
cells only weakly.\footnote{P1 is the mean-over-channels variance of the stream's first
difference, P2 the mean cosine of consecutive per-window gradients, and P3 static test-MSE over
static validation-MSE, read post hoc; none uses either online optimizer's trajectory. At the
default, P3 is the strongest single indicator (mean \GridPThreeSgdmWinMean{} in SGD-win against
\GridPThreeAdamWinMean{} in Adam-win cells), with P2 and P1 gaps of \GridPTwoGap{} and
\GridPOneGap. Drift-heavy, low-noise streams tolerate larger steps, so $10^{-3}$ happens to
remain usable on them.} At the default, these probes primarily describe where the shared rate
leaves Adam's empirical nonnegative-benefit range in this grid, rather than where either
optimizer is intrinsically preferable.

\subsection{C3 --- The Accuracy--Memory--Compute Frontier}\label{sec:frontier}

\begin{figure*}[t]
\centering
\includegraphics[width=\textwidth]{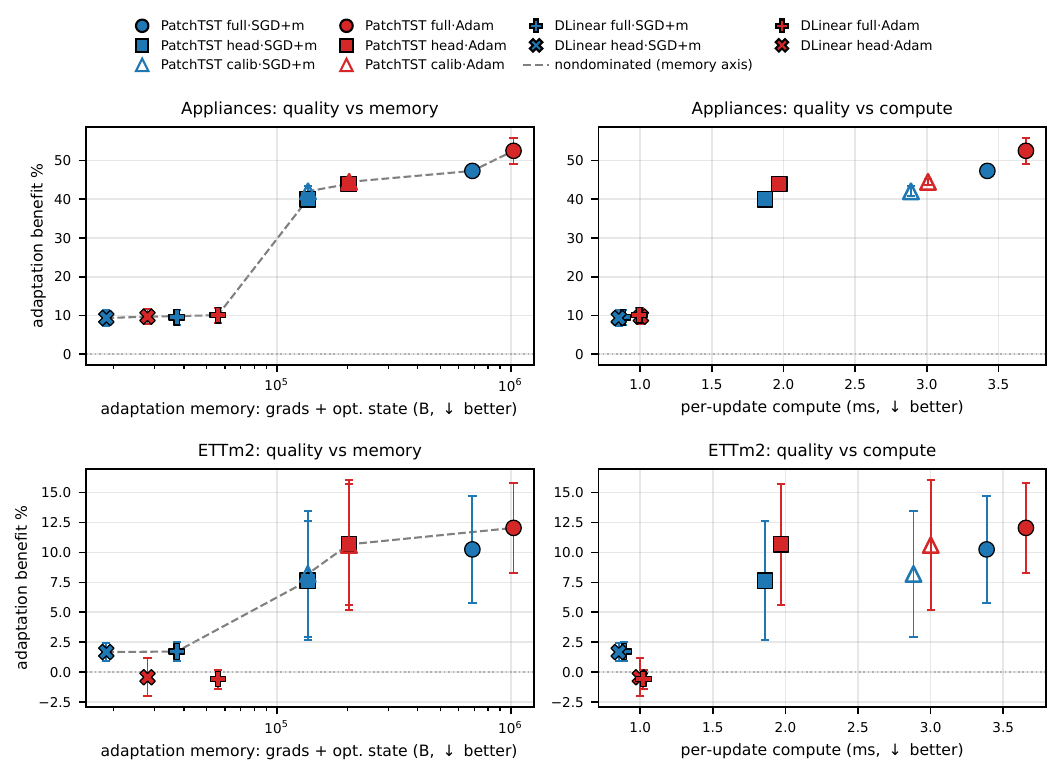}
\caption{Accuracy--resource characterization (C3) at validation-selected warmup and rehearsed learning
rates (five seeds; mean\,$\pm$\,standard deviation). Adaptation benefit is plotted against
adaptation-state memory (gradient buffers $+$ optimizer state; left) and A100-measured per-update
time (right) for Appliances (top) and ETTm2 (bottom). Color denotes optimizer (blue: SGD$+$m; red:
Adam); marker shape denotes backbone and parameter subset (PatchTST: circle/square/triangle for
full/head/calib; DLinear: plus/cross for full/head). Dashed links identify nondominated mean points
on the memory axis; hollow calib markers overlay head markers because their adaptation-state memory
differs only by the $2C$ calibration parameters.}
\label{fig:frontier}
\end{figure*}

Every frontier point uses the validation-selected warmup and rehearsed learning rate, both
chosen per dataset$\times$backbone$\times$strategy on the pre-drift validation slice. We place
each strategy on the accuracy-vs-resource plane over five seeds (mean$\pm$std,
Fig.~\ref{fig:frontier}); the full-model points coincide with the corresponding cells of
\S\ref{sec:lrfair}. The parameter-efficient strategies are evaluated on ETTm2 and Appliances,
the datasets with the smallest and largest oracle-reference benefits in
Table~\ref{tab:confound}.

\smallskip\noindent\textbf{Parameter-efficient trade-offs depend on the dataset.}
On PatchTST, \texttt{calib} (per-channel input/output calibration,
\FroEttmTwoPatchtstCalibSgdMParams{} trainable parameters,
$\approx$1/\FroEttmTwoFullOverCalibParams{} of the full model's) achieves a similar mean
improvement to full$\cdot$SGD+m on ETTm2
(\FroEttmTwoPatchtstCalibSgdMBenefit$\pm$\FroEttmTwoPatchtstCalibSgdMBenefitStd\% against
\FroEttmTwoPatchtstFullSgdMBenefit$\pm$\FroEttmTwoPatchtstFullSgdMBenefitStd\%), and
\texttt{head} matches it at the same adaptation-state memory but lower measured compute
(\FroEttmTwoPatchtstHeadSgdMMs\,ms against \FroEttmTwoPatchtstCalibSgdMMs\,ms: input-side
calibration forces a backward pass through the encoder, whereas head-only stops at the last
layer). Applying Adam to these subsets yields intermediate accuracy--memory trade-offs. On
Appliances, the six PatchTST points form a monotone mean trade-off: \texttt{calib}
\FroAppliancesPatchtstCalibSgdMBenefit\% at \FroAppliancesPatchtstCalibSgdMMemKb\,kB,
\texttt{calib$\cdot$Adam} \FroAppliancesPatchtstCalibAdamBenefit\% at
\FroAppliancesPatchtstCalibAdamMemKb\,kB, \texttt{full$\cdot$SGD+m}
\FroAppliancesPatchtstFullSgdMBenefit\% at \FroAppliancesPatchtstFullSgdMMemKb\,kB, and
\texttt{full$\cdot$Adam} \FroAppliancesPatchtstFullAdamBenefit\% at
\FroAppliancesPatchtstFullAdamMemKb\,kB. No plotted point dominates another in this setting.
On ETTm2, \texttt{head$\cdot$Adam} is nondominated relative to
\texttt{full$\cdot$SGD+m}: it obtains \FroEttmTwoPatchtstHeadAdamBenefit\% at
\FroEttmTwoPatchtstHeadAdamMemKb\,kB against \texttt{full$\cdot$SGD+m}'s
\FroEttmTwoPatchtstFullSgdMBenefit\% at \FroEttmTwoPatchtstFullSgdMMemKb\,kB. Thus,
whether PEFT$\times$Adam replaces full-model SGD$+$m or merely competes with it is
dataset-dependent. The cost of Adam's \emph{rate} and its \emph{state} are separable; in these
PatchTST settings, applying Adam to a smaller subset can retain a substantial fraction of the
observed full-model improvement at lower adaptation-state memory.

\smallskip\noindent\textbf{Rate sensitivity on the frontier.}
At the fixed default, the ETTm2 full$\cdot$Adam point reads
\FroEttmTwoPatchtstFullAdamBenefitFixed\% and appears dominated under that shared-rate
comparison; the rehearsed rate moves it to the highest-mean-benefit endpoint. Rehearsal is
mildly costly for full$\cdot$SGD+m: on Appliances it reads
\FroAppliancesPatchtstFullSgdMBenefit\% against \FroAppliancesPatchtstFullSgdMBenefitFixed\%
at its default. It leaves PEFT unchanged in this case, with \texttt{calib} at
\FroAppliancesPatchtstCalibSgdMBenefit\% and its rehearsal selecting the default rate in all
five seeds. The one sub-static Appliances/DLinear pick of \S\ref{sec:lrfair} does not appear
here because it uses a longer lookback and horizon than the frontier settings.

\smallskip\noindent\textbf{The adaptation-state-memory axis against an illustrative 192\,kB allowance.}
A resource axis is easiest to interpret against a budget. Federated split learning has been
demonstrated on smart meters carrying 192\,kB of SRAM~\cite{li2024edgemeter}. Relative to an
illustrative 192\,kB \emph{adaptation-state} allowance, full-model PatchTST adaptation exceeds
the allowance (\FroAppliancesPatchtstFullSgdMMemKb--\FroAppliancesPatchtstFullAdamMemKb\,kB of
gradients and optimizer state), PEFT with Adam is marginally above it
(\FroAppliancesPatchtstHeadAdamMemKb--\FroAppliancesPatchtstCalibAdamMemKb\,kB), and PEFT
with SGD$+$m is below it (\FroAppliancesPatchtstHeadSgdMMemKb--\FroAppliancesPatchtstCalibSgdMMemKb\,kB).
This comparison includes gradient buffers and optimizer state only; it excludes model weights,
activations, and runtime memory. It therefore illustrates how the adaptation-state accounting
changes the feasible strategy set, rather than establishing total-memory fit on a particular
meter.

\smallskip\noindent\textbf{A low update duty cycle under the A100 timing proxy.}
Under the A100 batch-1 timing proxy, one update per revealed horizon corresponds, at $H{=}24$
for an hourly meter, to one update per day. The \FroMsMin--\FroMsMax\,ms updates of
Fig.~\ref{fig:frontier} therefore amount to \FroSecPerYearMin--\FroSecPerYearMax\,s of
measured update time per meter per year. This duty-cycle calculation needs no power assumption;
only the equivalent \FroEnergyMinMj--\FroEnergyMaxMj\,mJ per update uses the 5\,W proxy. In the
two BDG2 measurements, per-update wall-clock is \ScBdgTwoSgdmMs/\ScBdgTwoAdamMs\,ms
(SGD$+$m/Adam) on the 15-meter subset and \ScBdgTwoRatAllSgdmMs/\ScBdgTwoRatAllAdamMs\,ms on
the 280-meter site. This is not an edge-device scaling result. The measurements suggest that
adaptation-state memory and the quality-vs-update-frequency (staleness) trade-off may be more
consequential than the measured A100 timing proxy.

\begin{figure*}[t]
\centering
\includegraphics[width=\textwidth]{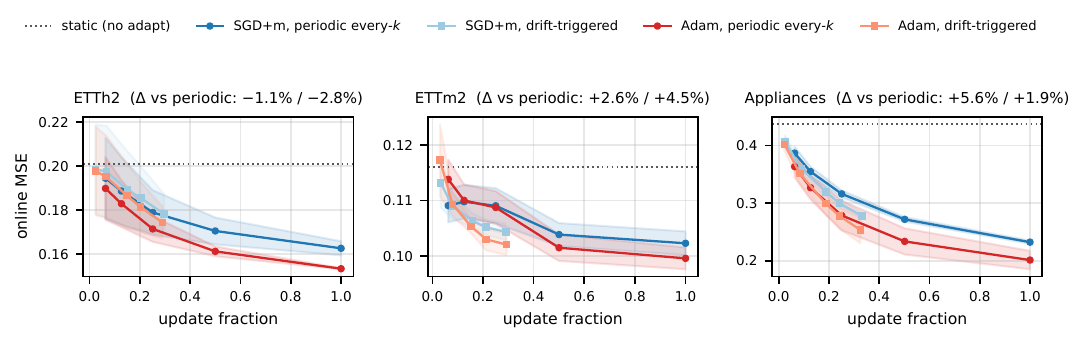}
\caption{Online MSE versus update fraction for PatchTST under validation-selected warmup and rehearsed
learning rates (three seeds; mean with $\pm$\,standard-deviation bands). Periodic every-$k$ and
drift-triggered schedules are shown for SGD$+$m and Adam in each dataset panel. Hue denotes
optimizer (blue: SGD$+$m; red: Adam), and the lighter shade denotes drift-triggered updating.
Panel-title values give the matched-budget difference between drift-triggered and periodic
schedules for SGD$+$m / Adam; positive values favor drift-triggered updating.}
\label{fig:staleness}
\end{figure*}

\smallskip\noindent\textbf{Staleness and drift-triggering.}
Quality improves with update frequency with diminishing returns (Fig.~\ref{fig:staleness}). At
validation-selected warmup and rehearsed rates, periodic and drift-triggered scheduling show
dataset- and optimizer-dependent differences (three seeds, mean$\pm$std). At matched update
budget, drift-triggering is ahead on both Appliances panels
(\StalAppliancesWinPct$\pm$\StalAppliancesWinPctStd\% lower MSE for SGD$+$m,
\StalAdamAppliancesWinPct$\pm$\StalAdamAppliancesWinPctStd\% for Adam) and both ETTm2 panels
(\StalEttmTwoWinPct$\pm$\StalEttmTwoWinPctStd\% and
\StalAdamEttmTwoWinPct$\pm$\StalAdamEttmTwoWinPctStd\%), and behind on both ETTh2 panels
(\StalEtthTwoWinPct$\pm$\StalEtthTwoWinPctStd\% and
\StalAdamEtthTwoWinPct$\pm$\StalAdamEtthTwoWinPctStd\%). No schedule dominates across the
examined datasets; schedule choice should therefore be validated per deployment.

\smallskip\noindent\textbf{Meter selection changes the estimated benefit.}
The 15-meter BDG2 subset of \S\ref{sec:setup} retains the least-missing meters of one site---a
choice that a 3,053-meter corpus makes possible to examine. Table~\ref{tab:bdgscale} reruns
the validation-selected protocol on five further subsets, selecting the optimizer separately
per seed by validation online MSE. On PatchTST, the \emph{anti-selected} 15 most-missing
meters of the same site report \MfiveBdgTwoRatWorstPatchtstSelBestMean\%, against
\MfiveBdgTwoPatchtstSelBestMean\% on the least-missing 15; on DLinear the ordering reverses
(\MfiveBdgTwoRatWorstDlinearSelBestMean\% against \MfiveBdgTwoDlinearSelBestMean\%). Thus,
fleet-scale estimates should report their meter-selection rule. The comparison is also affected
by what is being scored: in the anti-selected subset, many evaluation targets are forward-filled
rather than directly observed. Its larger reported benefit should therefore be interpreted as
performance on the imputed stream under this preprocessing rule, not as a data-quality-independent
estimate of adaptation benefit on observed readings. In particular, the anti-selected series have
a median of \DataBdgTwoRatWorstRepeatPct\% consecutive-equal samples, compared with
\DataBdgTwoRepeatPct\% for the \S\ref{sec:setup} subset. We introduced the inactive-meter rule
after observing a constant-value meter that flattened the fleet-wide metric; results using this
post-hoc data-quality rule should therefore be read as exploratory, although the rule uses only
warmup-pool statistics.

\smallskip\noindent\textbf{Additional scale within BDG2.}
With those rules fixed, the procedure is additionally evaluated on a 280-meter site, which
reports \MfiveBdgTwoRatAllPatchtstSelBestMean\%, and on a 240-meter, 18-site fleet, which
reports \MfiveBdgTwoFleetPatchtstSelBestMean\%. In these BDG2 subsets, the observed improvements
are higher for PatchTST than for DLinear. Every step---warmup selection, rate rehearsal, and
adaptation---remains a per-device computation.

\begin{table}[t]
\centering
\caption{\textbf{Meter selection and scale within BDG2 (three seeds; $H{=}24$, $L{=}96$).} Reported
values are adaptation benefit (\%) for the optimizer selected separately for each seed by
validation online MSE at its rehearsed rate. PatchTST entries are mean [min, max] across the
selected-seed results. ``Least/most-missing'' rank meters by missingness; ``active'' requires
coverage $\geq$50\% and the inactive-meter rule; the fleet retains up to the 15 least-missing
active meters per site (Swan has no meter meeting the coverage rule). The BDG2 row in
Table~\ref{tab:confound} is not directly comparable because it uses full$\cdot$SGD$+$m at the fixed
$10^{-3}$ rate and the test-selected oracle warmup.}
\label{tab:bdgscale}
\footnotesize\setlength{\tabcolsep}{4pt}
\begin{tabular}{l r c c}
\toprule
BDG2 subset & meters & PatchTST & DLinear \\
\midrule
Rat, least-missing 15 (\S\ref{sec:setup}) & 15 & \MfiveBdgTwoPatchtstSelBestMean{} [\MfiveBdgTwoPatchtstSelBestMin, \MfiveBdgTwoPatchtstSelBestMax] & \MfiveBdgTwoDlinearSelBestMean \\
Fox, least-missing 15 & 15 & \MfiveBdgTwoFoxPatchtstSelBestMean{} [\MfiveBdgTwoFoxPatchtstSelBestMin, \MfiveBdgTwoFoxPatchtstSelBestMax] & \MfiveBdgTwoFoxDlinearSelBestMean \\
Panther, least-missing 15 & 15 & \MfiveBdgTwoPantherPatchtstSelBestMean{} [\MfiveBdgTwoPantherPatchtstSelBestMin, \MfiveBdgTwoPantherPatchtstSelBestMax] & \MfiveBdgTwoPantherDlinearSelBestMean \\
Rat, \emph{most}-missing 15 (anti-sel.) & 15 & \MfiveBdgTwoRatWorstPatchtstSelBestMean{} [\MfiveBdgTwoRatWorstPatchtstSelBestMin, \MfiveBdgTwoRatWorstPatchtstSelBestMax] & \MfiveBdgTwoRatWorstDlinearSelBestMean \\
Rat, all active meters & 280 & \MfiveBdgTwoRatAllPatchtstSelBestMean{} [\MfiveBdgTwoRatAllPatchtstSelBestMin, \MfiveBdgTwoRatAllPatchtstSelBestMax] & \MfiveBdgTwoRatAllDlinearSelBestMean \\
Fleet: 18 of 19 sites & 240 & \MfiveBdgTwoFleetPatchtstSelBestMean{} [\MfiveBdgTwoFleetPatchtstSelBestMin, \MfiveBdgTwoFleetPatchtstSelBestMax] & \MfiveBdgTwoFleetDlinearSelBestMean \\
\bottomrule
\end{tabular}
\end{table}

\subsection{Conditional Deployment Guidance}\label{sec:recipe}
The measurements above provide conditional, per-device guidance without using test data.
\begin{itemize}
\item[\textbf{(i)}] Select the warmup budget \emph{and} online learning rate on the held-out
pre-drift validation slice at commissioning (\S\ref{sec:setup}).
\item[\textbf{(ii)}] When adaptation-state memory is the binding constraint, first consider
head-only or calibration-based adaptation. In the evaluated PatchTST settings, these variants
often retain a substantial fraction of the full-model gain at lower adaptation-state memory.
\item[\textbf{(iii)}] If the extra optimizer state and rehearsal are affordable, use Adam at
its rehearsed rate---1/10 to 1/30 of the default in \LrAdamSelTenthToThirtiethCells{} of the
\LrCells{} full-model cells---for a median \LrSelMedianGapPtAll\,pp (\S\ref{sec:lrfair}) at
$1.5\times$ the adaptation-state memory (\S\ref{sec:setup}). If state is what
binds, apply Adam to the calibration/head subset, or keep momentum alone.
\item[\textbf{(iv)}] If no tuning is possible, SGD$+$m at $10^{-3}$ was more rate-robust within
our grid; this observation should not be treated as a universal default. Our drift probe is post
hoc, so ``heavy drift'' is not decidable at commissioning.
\item[\textbf{(v)}] No schedule dominates across the evaluated datasets; validate periodic or
drift-triggered scheduling per deployment. The ETT/DLinear settings showed only small mean
improvements, so whether to adapt at all remains part of the decision.
\end{itemize}

\section{Discussion and Limitations}\label{sec:discussion}
\begin{itemize}
\item The validation-selection protocols spend the \emph{freshest} pre-drift data on selection
rather than training: the validation slice is never trained on, and there is no refit after
selection, so the deployed model's training data ends one validation span ($\approx$73 days on
the hourly and 15-minute sets, $\approx$14 days on Appliances) before the stream starts. Both
arms share this limitation, so the comparisons are internally consistent; however, a deployment
that folded the slice back into warmup training would use a fresher static baseline and could
change the reported benefit. Refit-after-selection, and the slice's size and position (fixed at
the adjacent 20\% throughout), remain unexplored.
\item The optimizer axis is deliberately the SGD$+$m/Adam pair, so the conclusion in
\S\ref{sec:lrfair} concerns \emph{that pair}, not optimizers in general. Whether memory-light,
learning-rate-free, and non-stationary-online optimizers (e.g., AdaFactor, Prodigy, and ObGD)
move the frontier points of \S\ref{sec:frontier} requires a separate study under the same
protocol. A direct comparison of our simplified calibration point with the official PETSA
implementation is likewise left for future work.
\item Everything is measured on a datacenter GPU. Adaptation memory is counted rather than
observed under an embedded allocator; per-update latency is launch-bound at batch~1 on an
A100, a regime an edge device does not share, so the near-flat scaling with meter count
(\S\ref{sec:frontier}) need not carry over; and the millijoule figures additionally assume a
5\,W device. Validating memory and latency on Jetson- or Raspberry-Pi-class hardware, and
ultimately on a meter, is required before the recipe can be called deployable end-to-end.
\end{itemize}

\section{Conclusion}
Conclusions about on-device time-series adaptation depend critically on the evaluation protocol.
Even under a leakage-free streaming protocol, we find that two additional evaluation choices
materially affect the estimated benefit: the baseline's \emph{warmup budget} (two-sided and
non-monotone on the seed-mean curves of our sweep) and a shared online \emph{learning-rate default}. Selecting the
warmup budget and each optimizer's rate on a held-out pre-drift validation slice mitigates,
rather than eliminates, these sensitivities without accessing test data. Under this
validation-only procedure, we characterize accuracy against adaptation-state memory and
A100-measured update latency on six public multivariate series, including a smart-meter corpus
at fleet scale. In the evaluated frontier settings, several parameter-efficient variants are
nondominated at lower adaptation-state memory than full-model alternatives. These measurements
provide conditional deployment guidance; target-device memory, latency, and energy remain to be
validated.

\section*{Acknowledgment}
This work was supported by the JSPS KAKENHI (Grant Number JP26K02884). The authors also
gratefully acknowledge support from the JST SIP Project (Grant Number JPJ012207).

\bibliographystyle{IEEEtran}
\bibliography{IEEEabrv,refs}

\end{document}